\documentclass[10pt,twocolumn,letterpaper]{article}

\usepackage[applications]{wacv}      

\usepackage{comment}

\newcommand{\Pyro}[0]{\textsc{PyroNear}$_{2025}$ }
\newcommand{\PyroV}[0]{\textsc{PyroNear}$_{2025}-\textsc{V}$ }
\newcommand{\PyroI}[0]{\textsc{PyroNear}$_{2025}-\textsc{I}$ }

\newcommand{\Pyrodate}[1]{\textsc{PyroNear}$_{#1}$ }

\usepackage{natbib}

\usepackage{multirow}

\definecolor{wacvblue}{rgb}{0.21,0.49,0.74}
\usepackage[pagebackref,breaklinks,colorlinks,allcolors=wacvblue]{hyperref}

\usepackage[dvipsnames]{xcolor}

\usepackage{multirow}
\usepackage{multicol}
\usepackage{pifont}
\usepackage{latexsym}
\newcommand{\cmark}{\textcolor{teal}{\ding{51}}}%
\newcommand{\xmark}{\textcolor{purple}{\ding{55}}}%

\def\wacvPaperID{67} 
\def\confName{WACV}
\def\confYear{2026}

\title{Constructing a Real-World Benchmark for Early Wildfire Detection with \\ the New \textsc{PyroNear}$_{2025}$ Dataset}

\author{
  Mateo Lostanlen\thanks{Core Contributors} \\
  PyroNear\\
  Paris, France \\
  {\tt\small mateo@pyronear.org} \\
  \and
  Nicolas Isla$^*$\\
  Universidad de Chile | CENIA\\
  Santiago, Chile \\
  {\tt\small nicolas.isla@ug.uchile.cl} \\
  \and
  Jose Guillen\\
  CENIA\\
  Santiago, Chile \\
  {\tt\small jose.guillen@cenia.cl} \\
  \and
  Renzo Zanca\\
  Universidad de Chile | CENIA\\
  Santiago, Chile \\
  {\tt\small renzo.zanca@ug.uchile.cl} \\
  \and
  Felix Veith\\
  PyroNear\\
  Paris, France \\
  {\tt\small felix@pyronear.org} \\
  \and
  Cristian Buc \\
  CENIA \\
  Santiago, Chile \\
  {\tt\small cristan.buc@cenia.cl} \\
  \and
  Valentin Barriere$^*$ \\
  DCC -- Universidad de Chile | CENIA\\
  Santiago, Chile \\
  {\tt\small vbarriere@dcc.uchile.cl}
}

\begin{document}
\maketitle 
\begin{abstract}
  Early wildfire detection (EWD) is of the utmost importance to enable rapid response efforts, and thus minimize the negative impacts of wildfire spreads. 
  To this end, we present \Pyro, a new dataset composed of both images and videos, allowing for the training and evaluation of smoke plume detection models, including sequential models. The data is sourced from: \textit{(i)} web-scraped videos of wildfires from public networks of cameras for wildfire detection in-the-wild, \text{(ii)} videos from our in-house network of cameras, and \textit{(iii)} a small portion of synthetic and real images. 
  This dataset includes around 150,000 manual annotations on 50,000 images, covering 640 wildfires, \Pyro surpasses existing datasets in size and diversity. It includes data from France, Spain, Chile and the United States. Finally, it is composed of both images and videos, allowing for the training and evaluation of smoke plume detection models, including sequential models. 
  We ran cross-dataset experiments using a lightweight state-of-the-art object detection model, as the ones used in-real-life, and found out the proposed dataset is particularly challenging, with F1 score of around 70\%, but more stable than existing datasets. Finally, its use in concordance with other public datasets helps to reach higher results overall. 
  Last but not least, the video part of the dataset can be used to train a lightweight sequential model, improving global recall while maintaining precision for earlier detections. 
   \textbf{We make both our code and data available online}.\footnote{\url{https://github.com/joseg20/wildfires2025}}    
\end{abstract}

\section{Introduction and Related Work}

With climate change, wildfire events are  increasing worldwide. Importantly, many of these devastating events emerge in remote areas that lack the infrastructure to implement solutions requiring on heavy computations and energy consumption. As a result, resource-efficient solutions have been explored to extend the applicability of wildfire detection systems, particularly in remote places that lack electrical power. \citet{DeVenancio2022} proposed an automatic fire detection system based on deep CNNs suitable for low-power, resource-constrained devices, achieving significant reductions in computational cost and memory consumption while maintaining performance. In the same vein, \citet{Khan2022} presented "FFireNet," a deep learning-based forest fire classification method, utilising a small neural network, the MobileNetV2 model for feature extraction, and achieving remarkable accuracy in binary classification of fire images. 

\paragraph*{Remote-sensing and wildfire detection}
Satellite imagery has been a pivotal data source for early wildfire detection. \citet{Barmpoutis2020} offered an overview of optical remote sensing technologies used in early fire warning systems. They conducted an extensive survey on flame and smoke detection algorithms employed by various systems, including terrestrial, airborne, and spaceborne-based systems. This review contributes to future research projects for the development of early warning fire systems. \citet{James2023} developed an efficient wildfire detection system utilizing satellite imagery and optimized convolutional neural networks (CNNs) for resource-constrained devices, using a MobileNet on an Arduino Nano 33 BLE. Whereas remote-sensing (and in particular satellite image-based methods) methods are particularly crucial to evaluate wildfire propagation across large areas, their temporal resolution prevents them to be optimal when it comes to detection speed, an issue where video-based detection presents strong advantages.

\paragraph*{Video-based fire detection}
These techniques have emerged as a promising avenue for early wildfire detection. \citet{Jin2023} provided a comprehensive review of deep learning-based video fire detection methods, summarizing recent advances in fire recognition, fire object detection, and fire segmentation using deep learning approaches. Their review provided insights into the development prospects of video-based wildfire detection for every kind of sequential images data, coming from various sources such as surveillance cameras, lookout towers, UAV or satellite sensors.

\citet{DeVenancio2023} proposed a hybrid method for fire detection based on spatial and temporal patterns, combining CNN-based visual pattern analysis with temporal dynamics to reduce false positives in fire detection. Additionally, \citet{Marjani2023} introduced "FirePred," a hybrid multi-temporal CNN model for wildfire spread prediction, emphasizing the importance of considering varying temporal resolutions in fire prediction models. Obviously, video-based early wildfire detection is strongly link to dataset quality, an issue we discuss next.

\paragraph*{Wildfire datasets}At a first glance, many of the datasets found in the literature could be useful for early wildfire detection. However, a deeper dive in those datasets show that they contain pictures of fires at an already advance stage, as exemplified in works such as \cite{Toulouse2017,Sharma17EANN,Foggia2015TCS}. In this work, where the accent is put over \textit{early} wildfire detection, we mainly focus on smoke plumes in order to detect early wildfires from watchtowers. As such, we discard the (easier) task of fire detection. 
In this context, it is notable to remark that only a very few of the datasets containing annotations for the smoke plume detection are publicly available.

In general, there are two main sources of videos for smoke plumes detection in the wild that are available online: \citet{HPWREN2023} (High Performance Wireless Research \& Education Network) and \citet{ALERTWildfire2023}. These two sources were used to create several datasets. 
Leveraging the camera network of the HPWREN, 
\citep{Dewangan2022,Govil2020,Mankind2023} propose annotated datasets for early wildfire detection, while other works \citep{Schaetzen2020,Yazdi2022,Mankind2023} propose datasets obtained from the ALERTWildfire network. 
Finally, from private sources and \textbf{not publicly available}, \citet{Fernandes2022} constructed a dataset of 35k images from Portugal that are annotated in smoke plumes. It is composed of 14,125 images that contain smoke plumes and 21,203 that do not.



\begin{figure*}[ht] \vspace*{-.5cm}
\centering
\includegraphics[width=1.\linewidth]{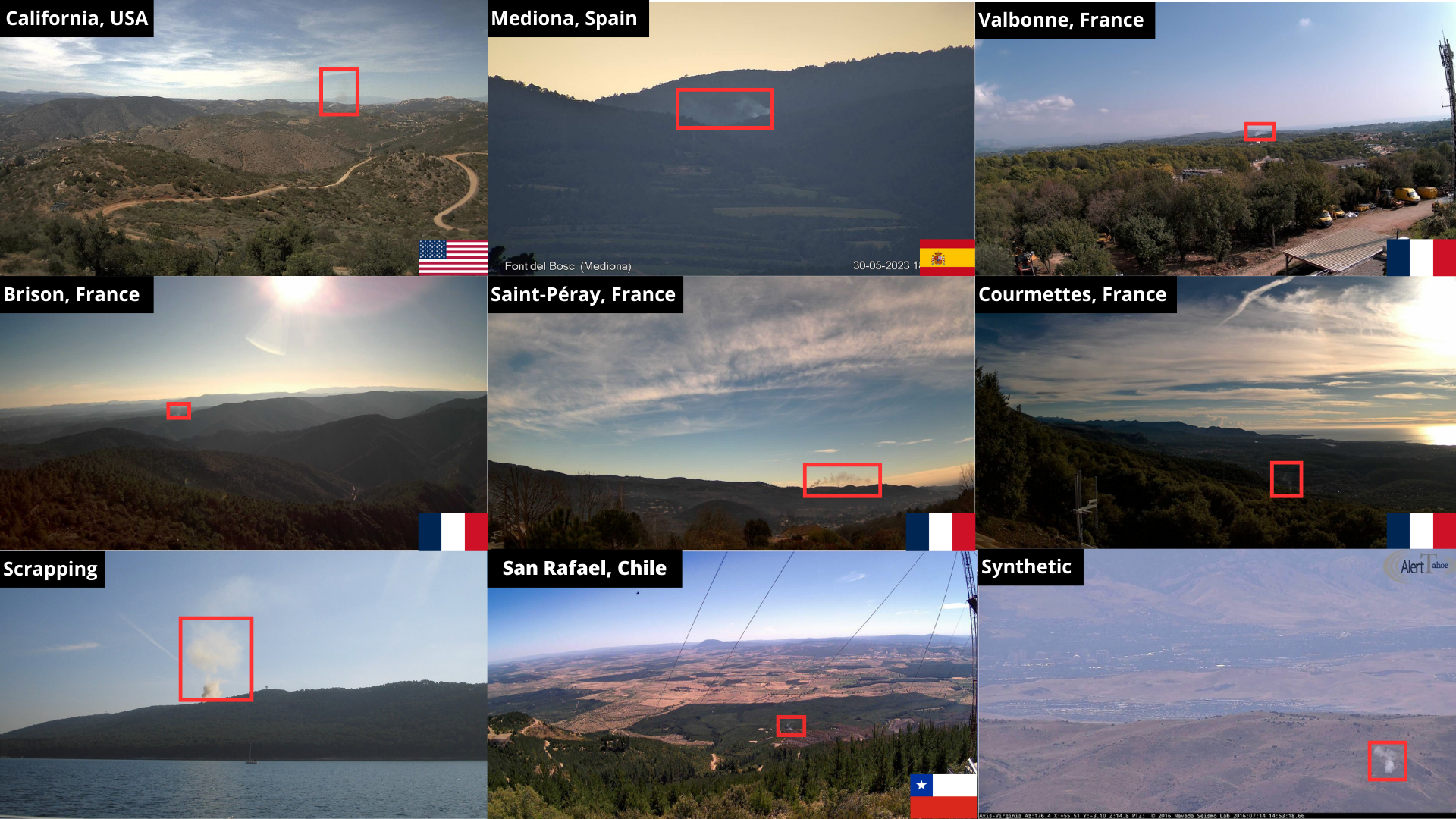} 
\caption{Examples of our dataset, containing real images and videos from France, Spain, United States and Chile, and synthetic images.}
\label{fig:pred_pyro} 
\end{figure*}

\paragraph*{Contributions} \textbf{\textit{This is a dataset resource paper}} for a concrete and useful application.\footnote{This is not a modelization paper.} Hence, we compare rigorously our dataset to existing ones, showing its interest: larger scale, greater diversity, increased complexity, support for sequential processing, and leading to better models. We provide simple baselines results, reflecting the real-world nature of the task: given the requirement for remote online data processing in environments without GPU availability, the system must remain lightweight. Consequently, the use of LLM, VLM, or video encoders is beyond the scope of our work. Snapshots of the dataset are visible in Figure \ref{fig:pred_pyro}.


\section{Datasets Collection, Fusion And Annotation}

\begin{figure*}
    \hspace*{-2.cm}
    \includegraphics[width=1.2\linewidth]{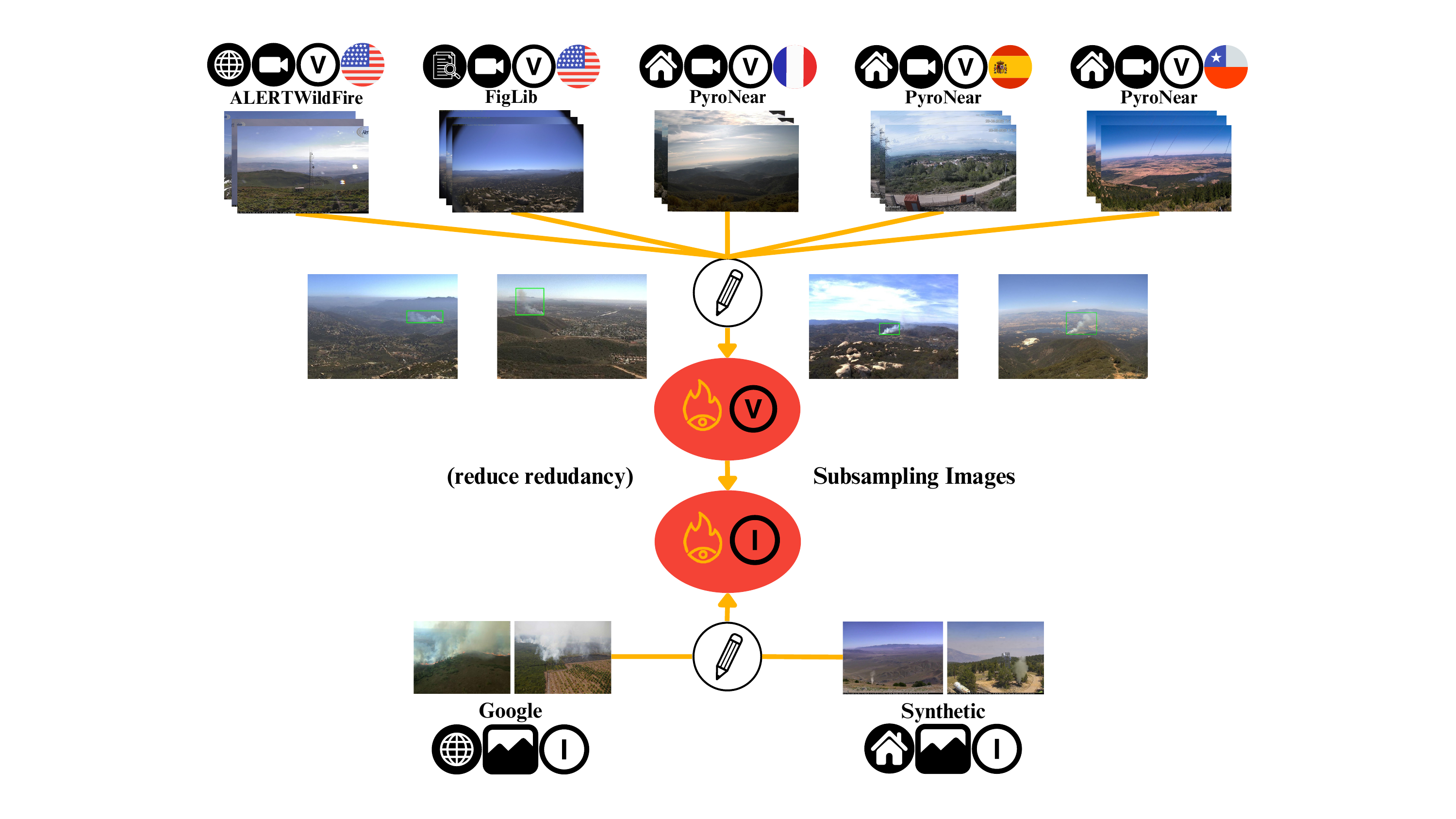} \vspace*{-1.cm}
    \caption{Summary of the whole process to create the video and the image datasets. ALERTWildfire data was collected from the web, FigLib data (without bounding box labels) comes from a published research paper, PyroNear data come from in-house data that we collected, Synthetic images were generated.}
    \label{fig:dataset_summary} \vspace*{-.6cm}
\end{figure*}

A summary of the whole process is visible in Figure \ref{fig:dataset_summary}. 

\subsection{Available Data}

In the development of an early wildfire detection model, the assembly of a comprehensive and diverse dataset is crucial. We already have a set of in-house data from our cameras in the wild, but in order to extend and diversify the dataset, we aimed at data from additional sources. This subsection outlines the primary sources of data and the derivative datasets coming from these sources, that have been annotated and widely used for past research on the topic. In this work, we will compare these datasets to our novel dataset. 

\paragraph*{Primary Data Sources}
Our data acquisition strategy leverages two main sources:

\begin{itemize}
    \item \textbf{\citet{HPWREN2023}:} Funded by the National Science Foundation, HPWREN is a non-commercial, high-performance, wide-area, wireless network of Pan-Tilt-Zoom (PTZ) cameras serving Southern California. It focuses on network research, including the demonstration and evaluation of its capabilities in wildfire detection.

    \item \textbf{\citet{ALERTWildfire2023}:} A consortium of universities in the western United States provides access to advanced PTZ fire cameras and tools, aiding firefighters and first responders in wildfire management, covering extensive regions spanning Washington, Oregon, Idaho, California, and Nevada. The ALERTWildfire website\footnote{\url{https://www.alertwildfire.org/}} grants public access to live feeds from these cameras. 
    
\end{itemize}

Note that Google is also used to collect images of wildfires, but this source contains only a small amount of smoke plumes images, as there are mainly close-up images of fire with big flames, which does not suit our purpose.

\paragraph*{Derived Datasets}
From these sources, several projects have proposed datasets that are of interest to our wildfire detection study:

\begin{itemize}
    \item \textbf{SmokeFrames:} Developed by \citet{Schaetzen2020} this dataset comprises nearly 50k images sourced from ALERTWildfire. To tailor it to our specific requirements of classical smoke plumes detection, we created a subset, \textit{SmokeFrames-2.4k}, consisting of 2410 images, from 677 different sequences, with an average of 3.6 images per sequence. The selected images were challenging for an in-house smoke plume detection model, triggering false positives. The original SmokeFrames-50k dataset was recreated by selecting 100 frames per video and removing images of nighttime fires to better align the dataset with our focus.

    \item \textbf{Nemo:} The dataset of \citet{Yazdi2022} includes frames extracted from raw videos of fires captured by ALERTWildfire's PTZ cameras, encompassing various stages of fire and smoke development.

    \item \textbf{Fuego:} Initiated by the Fuego project \citep{Govil2020}, this dataset was created by manually selecting and annotating images from the HPWREN camera network, based on historical fire records from Cal Fire. Out of 8500  annotated images with a focus on the early phases of fires, the authors make publicly available only a subset of 1661 images.

    \item \textbf{AiForMankind:} Two training datasets emerged from hackathons organized by AI For Mankind \citep{Mankind2023}, a nonprofit focusing on using AI for social good. These datasets, combined into one, offer a substantial collection of annotated images for smoke detection and segmentation.

    \item \textbf{FIgLib:} \citet{Dewangan2022} proposed the Fire Ignition image Library (FIgLib) composed of 24,800 images from South California from 315 different fires. It is the official dataset from the HPWREN. However, given that \textbf{the FigLig dataset does not initially have the bounding box annotations we are using in this work, we had to annotate it}.
    
\end{itemize}

Other datasets exist but they were unavailable due to proprietary restrictions \citep{Fernandes2022}, or efforts to obtain access through the original authors were unsuccessful \citep{Jeong2020}.
A summary of the existing datasets is visible in Table \ref{tab:dataset_summary}. 

\subsection{Creation of the \Pyro Dataset}

This section presents the collection of the data, its annotation using a homemade platform and a summary of the final dataset.  The global process is shown in Figure \ref{fig:dataset_summary} and snapshots of the data in Figure \ref{fig:pred_pyro}.

\subsubsection{Data Acquisition Strategy}


\paragraph*{Videos Web scrapping}

Our wildfire detection initiative utilizes the AlertWildfire camera network, which comprises approximately 130 cameras. The actual number of operational cameras fluctuates due to occasional unavailability, but 
despite these variances, we ensure comprehensive monitoring. 
The core of our data collection is an automated scraping script that interacts with the AlertWildfire API. 
This script retrieves images from each camera at the predetermined frequency of one image per minute, set by AlertWildfire. 
This gives a total of 
1,440 images per camera per day, summing up to about 187,200 images daily across the network.


\paragraph*{Videos Filtering}
After filtering out the night-time images, which are not of interest for this application, we perform inference on the remaining daylight images using a smoke plume detection model trained beforehand. This model analyzes each image, and any image with a wildfire detection score above 0.2 is marked as a potential fire event. To ensure comprehensive coverage of potential fire events, we also save images taken 15 minutes before and after each detection from the same camera. This approach helps capturing a broader contextual timeline around each potential wildfire incident, and collecting a dataset of videos in order to train and validate sequential models. 
All the images flagged during this process, including both potential wildfire detections and corresponding time-framed images, are stored for later annotation. This rich collection, encompassing potential early signs of wildfires as well as false positives, offers a challenging and valuable dataset for enhancing the performance 
in challenging scenarios that have historically led to a false detection. For example, 
in distinguishing true wildfires from non-threatening natural occurrences such as clouds, fog, or sunlight reflection. 

\paragraph*{In-house data from PyroNear cameras}

An in-house set of data was collected, using PyroNear stations\footnote{composed of cameras, a Raspberry Pi, and a 4G USB key} that were placed in 15 lookout towers in France, Spain, and Chile equipped with a total of 51 cameras.
The same process was performed using this network of cameras.



\paragraph*{Images Web scrapping}

This dataset was generated by scraping images from Google using keywords like "smoke" and "wildfire." After collecting the images, we manually filtered and selected 442 relevant images that depict various stages and types of smoke and wildfire. These are mean to provide a more diverse dataset, encompassing different environments and visual perspectives.

\paragraph*{Synthetic}

This dataset was created using images without fire, to which we added synthetic smoke plumes generated in Blender \citep{hess2013blender}. By applying Poisson blending, we randomly inserted these smoke plumes into the images, resulting in 200 images that mimic various smoke scenarios. This synthetic dataset helps enhance training for smoke detection models.

\begin{table*}[ht] 
\centering
{\resizebox{\textwidth}{!}{
    \begin{tabular}{@{}lccccccc@{}}
    \toprule
    Dataset          & Video   & \#Wildfires & Total Images$^*$ &  Train$^*$     & Validation$^*$ & Test$^*$ & Avg. BBox Area (\%) \\ 
    \midrule
    AiForMankind     & \xmark  & 31 & 2935/2584     & 2348/2042     & 294/263      & 293/279 & 1.341\\
    Fuego            & \xmark  & 38 & 1661/1572         & 1329/1269     & 166/152& 166/151    & 0.175  \\
    Nemo             & \xmark  & 62 & 2859/2570    & 2408/2158            & 251/237&    200/175 & 7.337
        \\
    SmokeFrames-2.4k & \xmark  & 75 & 2410/976       & 1928/781      & 241/125      & 241/70 &14.167
        \\
    SmokeFrames-50k & \cmark  & 643 & 54576/36304 & 45245/30282       & 4684/3213           & 4647/2809 & 14.384\\
    \hline \hline
    \PyroI           & \xmark  & 1049/640         & 4228/4041           & 3298/3161              & 489/462            & 441/418  &       1.976    \\
    \PyroV           & \cmark  &  1049/640          &  80727/55088      & 59283/40516          & 12206/8588              & 9238/5984 &              1.291        \\
    \bottomrule
    \end{tabular}
}}
\caption{Summary of Datasets. Columns marked with $^*$ indicate Total/Wildfire images. Our datasets have small bounding boxes, as we focus on EWD. One wildifre image generally contains one manually annotated bounding boxe, except SmokeFrames-50k which is semi-annotated and only contains 8,645 of them.}
\label{tab:dataset_summary} \vspace*{-.3cm}
\end{table*}

\subsubsection{Collaborative Annotation Platform}

\begin{figure}
    \centering
    \includegraphics[width=.45\textwidth]{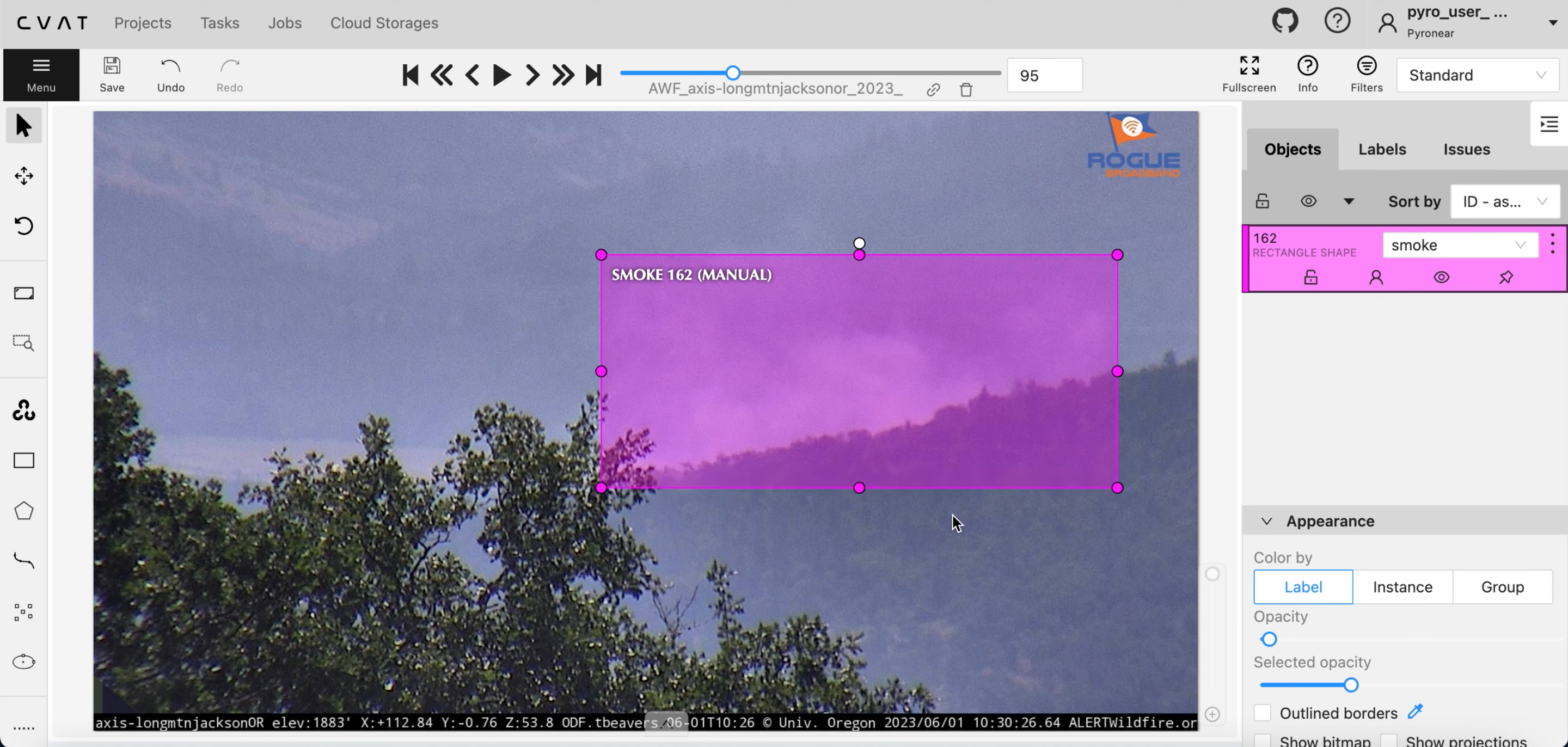}
    \caption{Snapshot of the Smoke Plume Annotation Platform.
    }
    \label{fig:platform} \vspace*{-.3cm}
\end{figure}

In order to annotate the wildfire data scrapped from the web, we developed a collaborative annotation tool with custom code in order to streamline the annotation process. In total, we collected a total of 150,000 annotations in a few month by leveraging the help of the PyroNear community. 
Platform usage was adapted and made as user-friendly as possible. This choice was motivated by the fact that annotators were all volunteers using their free time to help developing an open-source dataset and model. We gave the volunteers 150 images to annotate in order to maintain the annotation task short (less than 15 minutes), keeping the cognitive load low and allowing to avoid mistakes. Finally, we designed the platform in such a way that promotes a smooth and coherent workflow. 
A snapshot of the platform is visible in Figure \ref{fig:platform}. 

Initially, we started with an extensive collection of 120,000 annotations. With a 5-times cross-labeling approach, this pool was refined down to 24,000 unique images. 

Each of the images has been annotated by five annotators in order to minimize label errors. To validate the quality of the annotations, we calculated the inter-annotator agreement using \citet{Krippendorff2013}'s $\alpha$ with the presence or not of fire in each image, and obtained good agreement values. 

\vspace*{-.1cm}
\paragraph*{FigLib} 
Another contribution of this work is a new set of annotations on the 24,800 images of the FigLib dataset. Using the same annotation platform described above, we annotated every image with bounding boxes around the smoke plumes if present. 

\vspace*{-.2cm}
\paragraph*{Overall}
In the end, our dataset contains real images from United State, France, Spain and Chile collected over our own in-house network of cameras, the HPWREN and ALERTWildfire networks, and synthethic images we created, making the dataset very diverse and challenging. 
\subsubsection{Final Dataset: \Pyro}



The vast majority of the available datasets are not containing videos (see Table \ref{tab:dataset_summary}). Model training and performance assessment is done with classical object detection metrics, using independent images. In our case, we would like to emphasize that our dataset can be used in this setting, but also to train sequential models over video. For this reason, we chose to separate the data between a set that can be used to train and validate a model on sequence of images and the rest of the dataset. It gives us two datasets: \PyroI the one-image detection dataset, and \PyroV which contains the videos. The latter can be used to develop a temporal model that leverages a series of images for prediction, thus enhancing the accuracy and robustness of the detection. 
Both our image and video datasets focused on EWD, hence the average bounding box size is small compared to datasets like SmokeFrames or Nemo which contain data of PTZ camera after zooming with large bounding boxes. 


\paragraph*{\PyroI: Image dataset}
For the one-image object detection dataset, we discover that it is crucial to streamline the dataset to reduce redundancy affecting the model performance. 
Indeed, some events that are way longer contain many more images, leading to an unbalanced distribution with a lot of redundancy. In order to balance the dataset, we selected approximately 7 images per incident: one from the first detection, one without a fire, and the rest were randomly chosen to include images with fires. 
By retaining only 7 images per wildfire event, we effectively minimized repetitive or near-identical images. This approach was crucial in preserving the diversity of the dataset while ensuring its relevancy to our one-image object detection focus. 
The dataset was thus further refined to 4228 images, including 4041 smoke images. 
%
%
%
The final composition of the \Pyro dataset is visible in Table \ref{tab:dataset_summary}. 


\paragraph*{\PyroV: Video dataset}

The video part of the dataset contains 1049 videos of 640 different wildfires, from 4 different countries, making it the most diverse dataset of smoke plumes detection video. 
Snapshots of images from the dataset are available in Figure \ref{fig:pred_pyro}. 


\section{Experiments}

In this study, our primary objective is to evaluate the quality of various datasets by conducting a preliminary optimization process. 

\subsection{Single-Frame Methodology} \label{subsec:SF-methodo}


We use a small YOLOv8 model \citep{Yolov8}, renowned for its proficiency in diverse detection scenarios. Given the nature of our task, and the necessity for frugal computing, the small version of the model was chosen for its balance between speed, size and accuracy. 
The optimal batch size and number of epochs were found using a grid search in $\{50,100\}$ and $\{2^k, k=5, 6\}$ on the validation set. 
Alongside this, we also identify the optimal confidence threshold $\tau_d$ in \{$k.10^{-2}, k=1...20$\} in the same way. 



\paragraph*{Dataset Splitting Strategy}

To prepare our datasets for model training and validation, we followed the existing split in the Nemo dataset, where approximately 9.3\% of the data was allocated for validation. To maintain consistency across all datasets and ensure a comparable evaluation framework, we adopted a similar approach for the other datasets, targeting a close approximation of a 10\% split for the validation set, while also ensuring that another 10\% is allocated for the test set. This strategy enables a balanced and uniform methodology for assessing the performance of our models across different datasets, ensuring that each dataset is represented fairly in both training, validation and testing phases. Finally, and in order to maintain independent partitions, we kept the wildfire events disjoints between the train, val and test. This was not possible for datasets like Nemo or SmokeFrames, as the files were named with different names, from the same wildfire but different perspectives.
%
Nevertheless, due to conflicts arising from overlapping images between the Nemo and SmokeFrames datasets, we filtered the problematic images from SmokeFrames test sets. We used perceptual hash\footnote{\url{https://github.com/knjcode/imgdupes}} along with Hamming distance to ensure that duplicate or highly similar images were excluded from SmokeFrames test set.

\paragraph*{Metrics}

Following past works \citep{Schaetzen2020,Yazdi2022,Dewangan2022} we use precision, recall, and the F1 score as metrics in order to validate the different models. We chose not to use the usual object detection metric such as mean average precision (mAP) as 
the goal is about correctly classifying areas in an image as indicating the presence or not of a wildfire, without being able to get the contours of the smoke plumes which can be subjective. 



\subsection{Video-Based Methodology}
Inspired by the work of \cite{Jeong2020}, we employed a modified approach where a smoke plume detector was employed to extract bounding boxes. The coordinates of the bounding boxes allows for detecting an area of interest in the image, which was then processed by a pre-trained ResNet \citep{he2015deepresiduallearningimage} in order to extract a sequence of learned representations. 
Finally, to process temporal information, we employed a simple LSTM for binary classification
, consistent with the previous work's methodology.

\section{Baselines Results}

\subsection{Image Dataset Evaluation}

\begin{figure*}[ht]
\centering
\includegraphics[width=1.\linewidth]{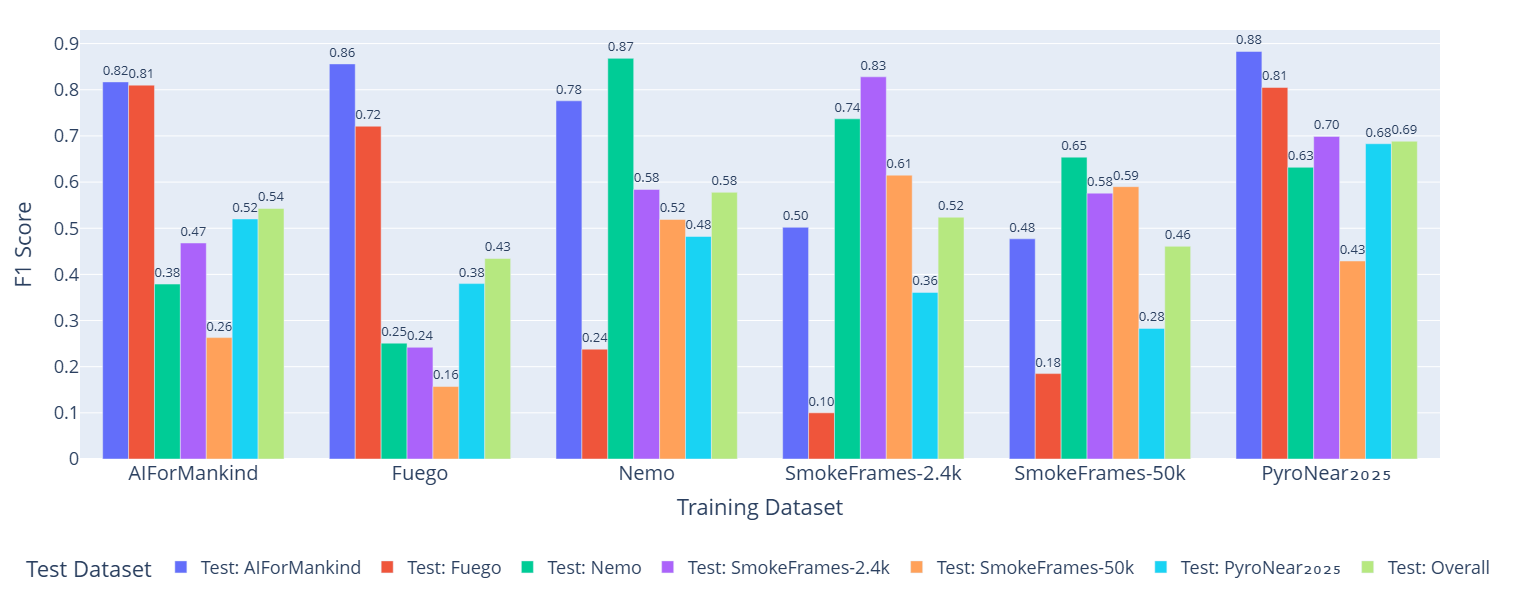} \vspace*{-.4cm}
\caption{F1 scores obtained by training on each dataset (x-axis) and evaluating on all others, including self-evaluation.} \vspace*{-.4cm} 
\label{fig:cross-validation}
\end{figure*}

The test set results of the smoke plumes detection models with their associated threshold are shown in Figure Table \ref{tab:best_model_per_dataset_details}. The datasets with the lower F1 scores are SmokeFrames-50k, making them apriori the most challenging ones. We will confirm in the following section that only our dataset is challenging. 
It is notable that the optimum detection threshold plays an important role as it can vary up to 5 times in size, going from 0.04 to 0.19. 




\begin{table}[ht]
\centering
\begin{tabular}{@{}llc@{}}
\toprule
Dataset            & F1 Score & $\tau_d$ 
\\ \midrule
SmokeFrames-2.4k      &     0.828   &     0.19   
\\
SmokeFrames-50k      &     0.576   &     0.05  
\\
Nemo                    &   0.868               &   0.09        
\\
AiForMankind    &    0.817              &     0.09   
\\
Fuego                   &   0.721               &  0.04       
\\
\PyroI               &     0.683             &    0.11

\\ \bottomrule
\end{tabular}
\caption{Results of the best performing models for each dataset on the associated test split, with the optimal detection threshold $\tau_d$.
}
\label{tab:best_model_per_dataset_details} \vspace*{-.3cm}
\end{table}

\subsubsection{Cross-Dataset Model Evaluation}


%
The performances of the models trained on a cross-dataset setting dataset on the different test sets are shown in Table \ref{tab:cross_datasets}. 
The results over the different datasets are very variable, but there is a clear trend suggesting that models obtain the best test results when they have been trained with elements from the same dataset. This is especially true for Nemo and SmokeFrames-2.4k that reach F1-scores of 86.8\% and 82.8\% on their respective test sets. 
However, \PyroI allows reaching the best results overall (F1 score of 68.8\%). For this reason, we believe that the high performances of Nemo and SmokeFrames are mainly due to overfitting issues because of the partitioning (see Section \ref{subsec:SF-methodo}). 
The bounding boxes average size also plays a role, as the dataset with similar bounding box size\footnote{Fuego/AI4Mankind/\PyroI $\approx$1\% vs Nemo/SmokeFrames-2.4k/SmokeFrames-50k $\approx$ 10\%} works better between them.

Furthermore, we trained a model on the combined dataset, joining the train, validation and test sets respectively together, and display the results in Table \ref{tab:combine_performance_model}. The two datasets that remain challenging, with an F1 lower than .9, are \PyroI and SmokeFrames-50k. We believe the latter remains difficult, even though with large bounding boxes, because of noisy annotations coming from a semi-supervised process without human validation. 
Indeed, when looking at the prediction of our model on the dataset, we found out it detects a smoke plume at the right place before it gets a real annotation (more details in Appendix, Section \ref{app:early_detection}). 

\begin{table}[ht]
\centering
\begin{tabular}{@{}lccc@{}}
\toprule
Tested Dataset      & Precision & Recall & F1 Score \\ \midrule
AiForMankind        &    0.916 & 0.903 & 0.910     \\
Fuego               &     0.959 & 0.940 & 0.950    \\
Nemo                &    0.956 & 0.915 & 0.935    \\
SmokeFrames-2.4k    &     0.824 & 0.986 & 0.897    \\
SmokeFrames-50k    &     0.787 & 0.531 & 0.634   \\
\PyroI               &     0.838 & 0.745 & 0.789    \\ \hline
Overall             &     0.880     &   0.836     &     0.852     \\
\bottomrule
\end{tabular}
\caption{Model performance when trained over a combined dataset.}
\label{tab:combine_performance_model} \vspace*{-.2cm}
\end{table}

\begin{figure*}[!ht]
    \centering
    \begin{subfigure}[b]{0.45\textwidth}
        \centering
        \includegraphics[width=.9\textwidth]{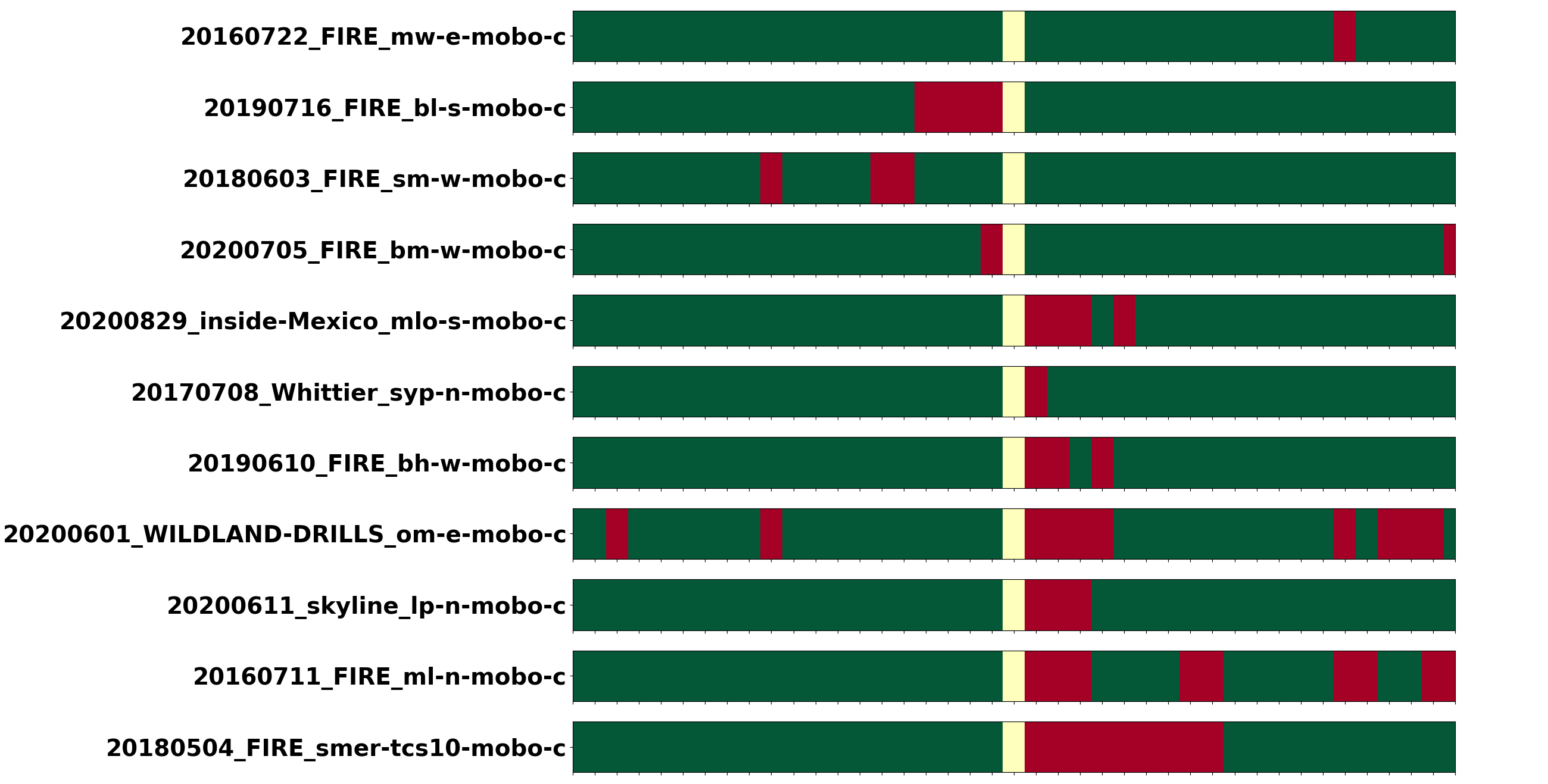} 
        \caption{Single-frame model prediction compared to the ground-truth}
    \end{subfigure}
    \hspace{0.05\textwidth} 
    \begin{subfigure}[b]{0.45\textwidth}
        \centering
        \includegraphics[width=.9\textwidth]{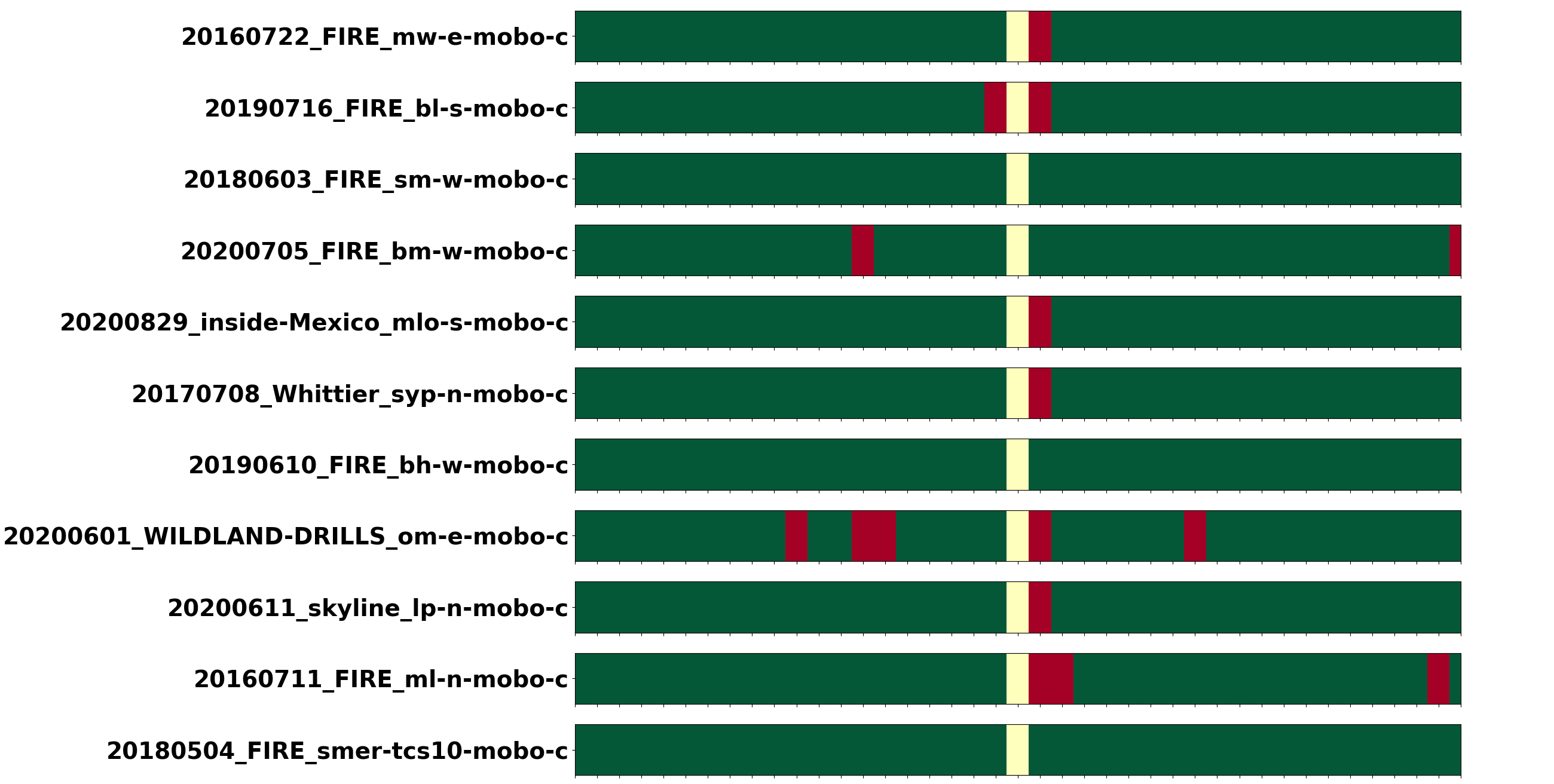} 
        \caption{Video model prediction compared to the ground-truth}
    \end{subfigure}
    \caption{Comparison of the predictions of a single-frame model versus a video model on a set of videos from FigLib.} \label{fig:video_prediction_small} \vspace*{-.4cm}
\end{figure*}

\subsubsection{Threshold Sensitivity Analysis}

\begin{figure}[ht]
\centering
\includegraphics[width=\linewidth]
{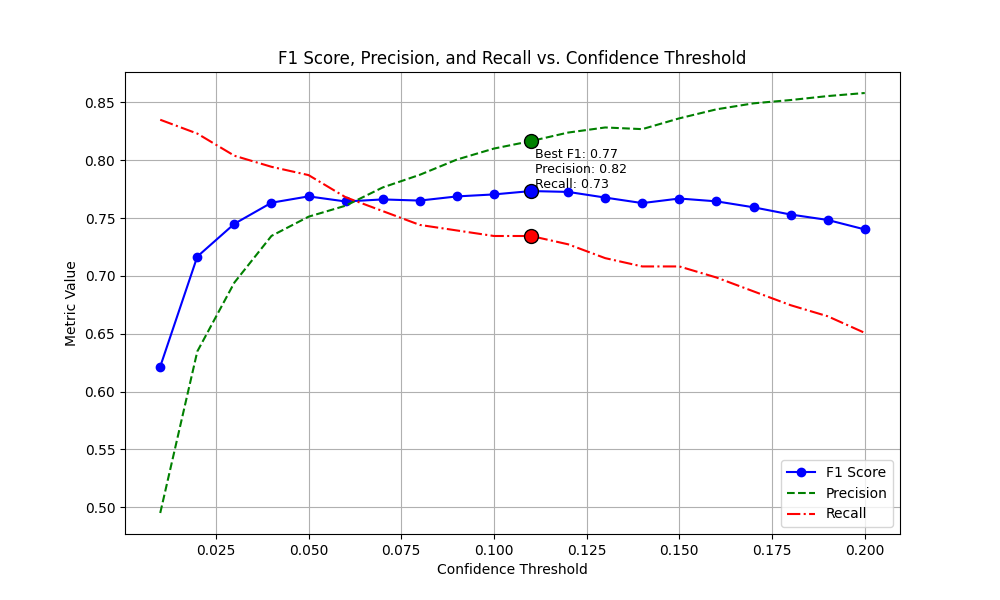} 
\caption{F1-score on the validation split of \PyroI w.r.t. detection threshold $\tau_d$}
\label{fig:f1_curves} \vspace*{-.3cm}
\end{figure}

Curves in figure \ref{fig:f1_curves} represent the performance of the models across different detection thresholds, which is an hyperparameter representing the minimum values to consider a detection as a positive. The curves highlight the trade-offs between precision and recall at various thresholds, showing that recall (an important metric that reflect early plume detection) can be higher as a function of this threshold. The latter is then set up to a lower value (with more false positives) in the multi-frame detection setting, as the sequential model helps diminishing the number of false positives. 

\subsubsection{Synthetic Images Ablation}

We ran the experiments with our dataset without the synthetic images, in order to show their interest. Without them, the performances are dropping a bit, with an overall F1-score 2.0\% lower on the test part of the \PyroI dataset. 

\subsection{Video Dataset Evaluation}

Finally, we propose a baseline for the video part of our dataset, aiming for models that detect better and sooner the smoke plumes. Results are visible in Table \ref{tab:PyroV_results}, comparing an image-base model with a sequential model. As the sequential model utilizes the detection of the first model with a very low detection threshold and process them again in a sequential way, it helps to increase the general recall without hurting the precision (10.065\% versus -1.491\% in relative). Moreover, the sequential approach allows to reduce the necessary time to detect the fire from 35 seconds in average. In Figure \ref{fig:video_prediction_small} are compared on a sample of examples videos at the frame-level if predictions from the Vanilla model and the sequential model are true or false. The model sequential model trained with our video dataset allows for both earlier detections and for smoothing the predictions.



\begin{table}[ht]
\centering
\resizebox{\linewidth}{!}{%
\begin{tabular}{@{}lcccc@{}}
\toprule
Model      & Precision & Recall & F1 Score & Time Elapsed \\ \midrule
Vanilla (1 frame)      &   \textbf{0.805}       &  0.775      &     0.790    & 1.76 \\ 
Sequential         &   0.793        &  \textbf{0.853}      & \textbf{0.822}     &   \textbf{1.17} \\ 
\bottomrule
\end{tabular}
}
\caption{Performance comparison of image-based and sequential models on the \PyroV dataset using Precision, Recall and F1, as well as the time elapsed (mn) before detecting the fire.}
\label{tab:PyroV_results} 
\vspace*{-.3cm}
\end{table}

\section{Conclusion}

In this paper we presented \Pyro, a new dataset for smoke plume detection. We collected it by scrapping online data and using an already trained model in order to filter out challenging examples such as detection and false positives. 
The dataset was then re-annotated by a pool of volunteers using an online platform designed for the purpose. We merged it with new in-house data, and existing non annotated datasets, making our dataset the most diverse open-source for this domain, with data from four different countries. 
We showed that our dataset is challenging, with small size smoke plumes for real-life early wildfire detection, and that training using our dataset helps to globally improve smoke plume detection models on other public datasets. 
Finally, We kept the images before and after every fire event detection in order to generate videos. This allows us to propose a video part of the data, allowing to train sequential models, which improve the global recall over classical single-frame smoke plume detection models while keeping similar precision. This data collection and annotation effort will be pursued in order to extend this dataset to other domains, such as new landscape and meteorological conditions, and it will be put online for research and non-profit purposes. 

\section*{Limitations and Future Work}

While the cross-labeling approach used for \Pyro has significantly contributed to the accuracy of our dataset, it has also led to a substantial reduction in the number of images we could include. Acknowledging this limitation, we are currently developing a new methodology for the upcoming \Pyrodate{2026} dataset, which aims to semi-annotation process.

\vspace*{-.2cm}
\paragraph*{Faster Annotation} We are exploring semi-automatic annotation techniques that will accelerate the labeling process while maintaining high-quality annotations. By integrating advanced algorithms with manual oversight, we can swiftly annotate large volumes of images without compromising on accuracy.

\vspace*{-.2cm}
\paragraph*{Normalization of Annotations} The semi-automatic approach also aims to standardize the annotation process across different users. This consistency is crucial for ensuring that the dataset reflects a uniform understanding of wildfire and smoke characteristics.

\vspace*{-.2cm}
\paragraph*{Reduced Cross-Labeling} With the improved efficiency and consistency brought by semi-automatic annotation, we anticipate the need for cross-labeling to decrease significantly. This reduction will enable us to retain a larger portion of the images initially collected, thereby enriching the \Pyrodate{2025} dataset with a broader range of data.

These advancements are expected to not only enhance the volume of annotated data but also to improve the overall quality and representativeness of the \Pyrodate{2025} dataset. 

\bibliographystyle{plainnat}
\bibliography{JRC_mendeley,Wildfire_detection}

\clearpage
\setcounter{page}{1}
\maketitlesupplementary
\setcounter{section}{0}


\section{Full Results} \label{app:full_res}

The full results, including the Precision, Recall and F1-scores of each models on each test set are visible in Table \ref{tab:cross_datasets}. 

   \begin{table*}[ht]
\centering
\resizebox{.6\textwidth}{!}{
\begin{tabular}{@{}llccc@{}}
\toprule
Train Dataset      & Test Dataset      & Precision & Recall & F1 Score \\ \midrule

\multirow{7}{*}{\PyroI} 
& \textbf{AIForMankind} & 0.876 & 0.889 & \textbf{0.883} \\
& Fuego      & 0.930 & 0.709 & 0.805 \\
& Nemo                & 0.738 & 0.553 & 0.632 \\
& SmokeFrames-2.4k    & 0.611 & 0.817 & 0.699 \\
& SmokeFrames-50k     & 0.572 & 0.343 & 0.429 \\
& \textbf{\Pyro}     & 0.736 & 0.637 & \textbf{0.683} \\ \cline{2-5}
& \textbf{Overall}             & \textbf{0.743} & \textbf{0.658} & \textbf{0.688} \\ \hline


\multirow{7}{*}{Nemo} 
& AiForMankind        & 0.983 & 0.641 & 0.776 \\
& Fuego               & 0.840 & 0.139 & 0.238 \\
& \textbf{Nemo}       & 0.876 & 0.860 & \textbf{0.868} \\
& SmokeFrames-2.4k    & 0.415 & 0.983 & 0.584 \\
& SmokeFrames-50k     & 0.587 & 0.466 & 0.519 \\
& \Pyro                 & 0.670 & 0.376 & 0.482 \\
\cline{2-5} 
& Overall             & \textbf{0.729} & \textbf{0.578} & \textbf{0.578} \\ \hline

\multirow{7}{*}{AiForMankind} 
& AiForMankind        & 0.852 & 0.784 & 0.817 \\
& Fuego               & 0.902 & 0.735 & 0.810 \\
& Nemo                & 0.607 & 0.275 & 0.379 \\
& SmokeFrames-2.4k    & 0.550 & 0.550 & 0.468 \\
& SmokeFrames-50k     & 0.346 & 0.211 & 0.263 \\
& \Pyro               & 0.558 & 0.487 & 0.520 \\ \cline{2-5}
& Overall             & \textbf{0.636} & \textbf{0.507} & \textbf{0.543} \\ \hline

\multirow{7}{*}{SmokeFrames-2.4k} 
& AiForMankind        & 0.989 & 0.336 & 0.502 \\
& Fuego               & 1.000 & 0.050 & 0.100 \\
& Nemo                & 0.902 & 0.623 & 0.737 \\
& \textbf{SmokeFrames-2.4k} & 0.779 & 0.883 & \textbf{0.828} \\
& \textbf{SmokeFrames-50k}     & 0.777 & 0.508 & \textbf{0.615} \\
& \Pyro               & 0.727 & 0.240 & 0.361 \\ \cline{2-5}
& Overall             & \textbf{0.862} & \textbf{0.440} & \textbf{0.524} \\ \hline

\multirow{7}{*}{SmokeFrames-50k} 
& AiForMankind        & 0.612 & 0.390 & 0.477 \\
& Fuego               & 0.351 & 0.125 & 0.185 \\
& Nemo                & 0.670 & 0.639 & 0.654 \\
& SmokeFrames-2.4k    & 0.445 & 0.816 & 0.576 \\
& SmokeFrames-50k     & 0.697 & 0.511 & 0.590 \\
& \Pyro               & 0.352 & 0.236 & 0.283 \\ \cline{2-5}
& Overall             & \textbf{0.521} & \textbf{0.453} & \textbf{0.461} \\ \hline

\multirow{7}{*}{Fuego} 
& AiForMankind        & 0.817 & 0.899 & \textbf{0.856} \\
& Fuego               & 0.782 & 0.668 & 0.721 \\
& Nemo                & 0.342 & 0.198 & 0.251 \\
& SmokeFrames-2.4k    & 0.212 & 0.283 & 0.242 \\
& SmokeFrames-50k     & 0.220 & 0.121 & 0.157 \\
& \Pyro               & 0.404 & 0.359 & 0.380 \\ \cline{2-5}
& Overall             & \textbf{0.463} & \textbf{0.421} & \textbf{0.435} \\ \hline

\bottomrule
\end{tabular}
}
\caption{Results of the cross-over experiment using the test sets of each dataset. In bold, the best results for each combination. The “Overall” section contains the averages for each metric. SmokeFrames-2.4k has been re-annotated in this work.} \vspace*{-.4cm}
\label{tab:cross_datasets}
\end{table*}

\section{Results with the different hyperparameters} \label{app:results_HP}


We provide below the 1 score curves for each dataset. 
These curves, shown below each table, visually represent the performance of the models across different confidence thresholds. The confidence threshold represent the minimum values for which we consider the detection as a positive. The inclusion of F1 curves offers an intuitive understanding of the model's classification performance, highlighting the trade-offs between precision and recall at various thresholds. 

\begin{figure}[ht]
\centering

\begin{subfigure}[b]{0.48\linewidth}
    \includegraphics[width=\linewidth]{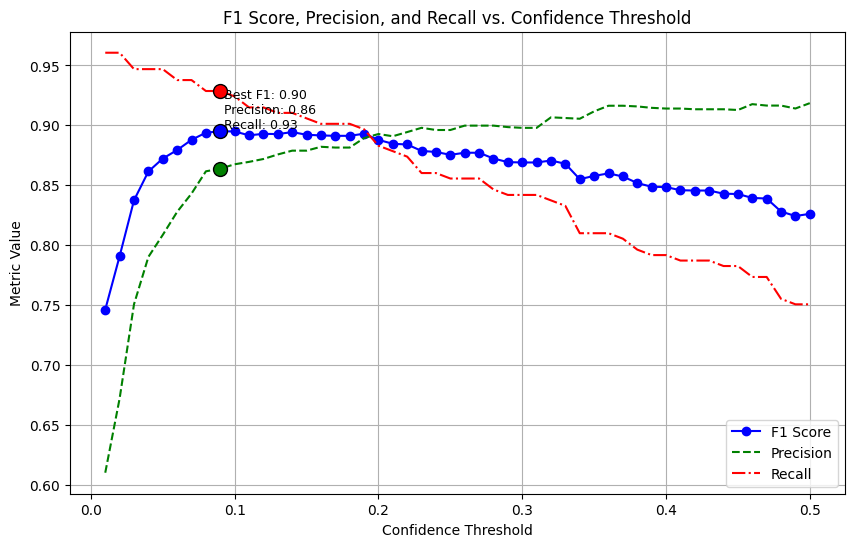}
    \caption{Nemo}
    \label{fig:f1_curves_nemo}
\end{subfigure}
\hfill
\begin{subfigure}[b]{0.48\linewidth}
    \includegraphics[width=\linewidth]{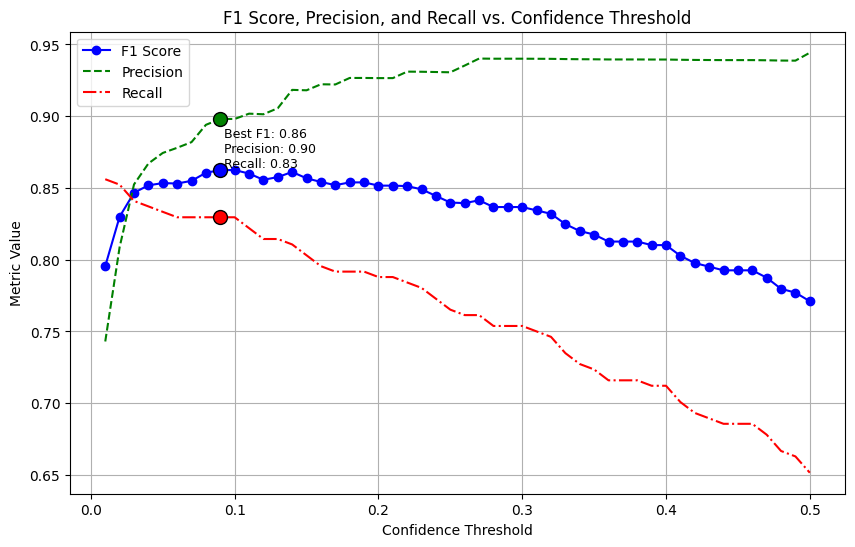}
    \caption{SF-2.4k}
    \label{fig:f1_curves_sf}
\end{subfigure}

\vspace{1em}

\begin{subfigure}[b]{0.48\linewidth}
    \includegraphics[width=\linewidth]{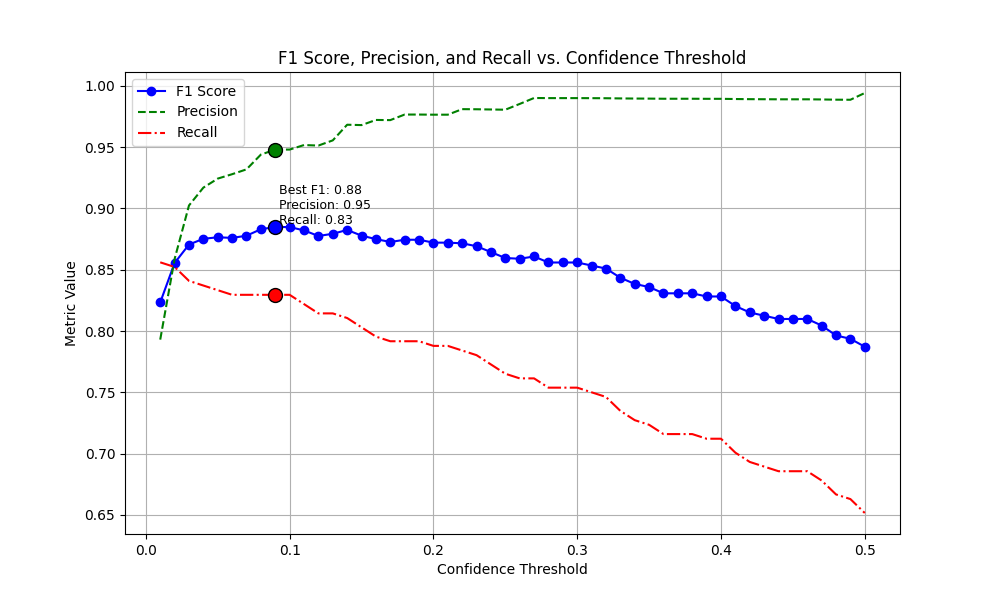}
    \caption{AI4M}
    \label{fig:f1_curves_ai}
\end{subfigure}
\hfill
\begin{subfigure}[b]{0.48\linewidth}
    \includegraphics[width=\linewidth]{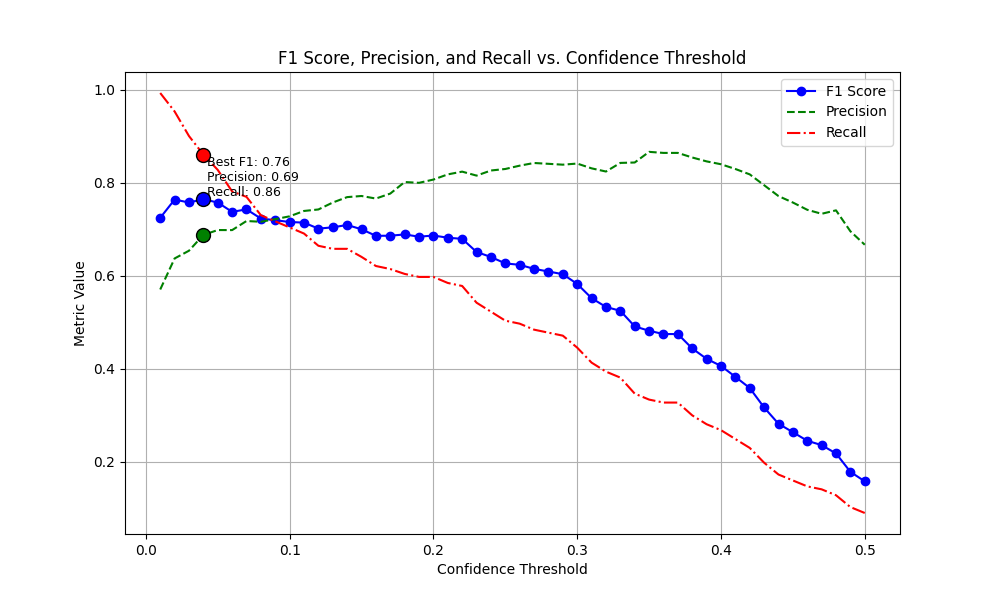}
    \caption{Fuego}
    \label{fig:f1_curves_fuego}
\end{subfigure}

\caption{F1 Score at various thresholds across datasets on their validation splits: Nemo, Fuego, AI4M, and SF-2.4k.}
\label{fig:f1_curves_all}
\end{figure}

\section{Sequential model Methodology}

\begin{figure*}
\centering
    \includegraphics[width=1.\textwidth]{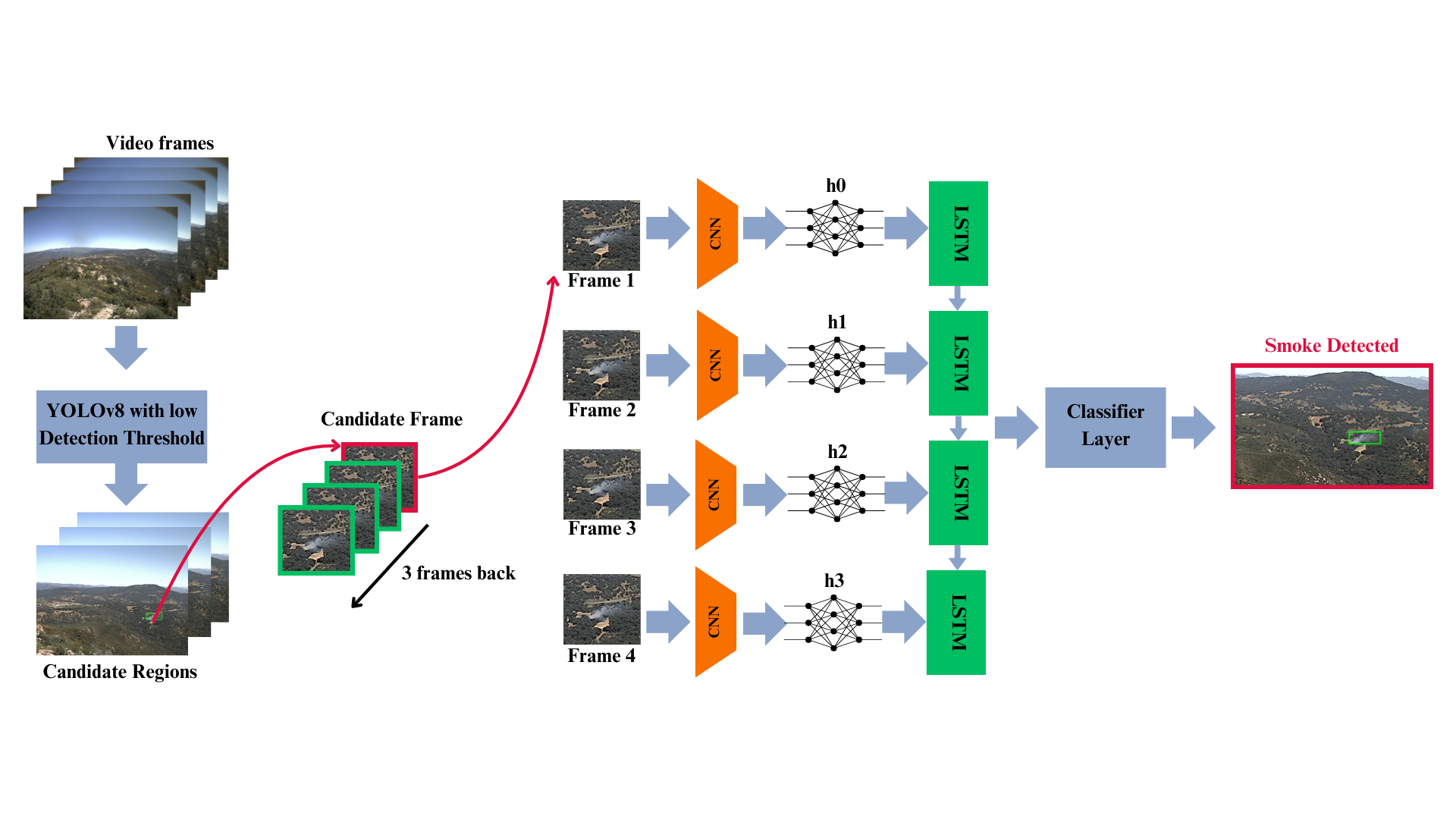}
    \caption{Architecture of the sequential model, which used the last four frames as context windows in order to detect if the patch contains a smoke plume in a binary way.}
    \label{fig:cnnlstm}
\end{figure*}

We based the architecture of our sequential model on \cite{Jeong2020}, without the knowledge distillation. The CNN used is a pre-trained Resnet50, followed by an LSTM. The weights of the YOLOv8 are frozen but the detection threshold $\tau_d$ is set as a new hyperparameter. The best model was found with $\tau_d=10^{-2}$, and one LSTM layer with a hidden size of 256, and a ResNet50 as image encoder. Figure \ref{fig:cnnlstm} shows more details on the architecture. 

For the training of the sequential model, we samples sequences of images in order to get a mix of both positive and negative examples. For this, we selected sequences that were before the beginning of the fire, when the fire had already started, and when the fire was starting (as our goal is EWD). Finally, we also add the sequence of examples that were detected as false positive with the single-frame model, using the same lower confidence threshold used afterward.\footnote{As we reduced this threshold to increase the global Recall.}

\section{Time to detect a wildfire}

\begin{figure*}[htbp]
    \centering
    \begin{subfigure}[b]{0.45\textwidth}
        \centering
        \includegraphics[width=\textwidth]{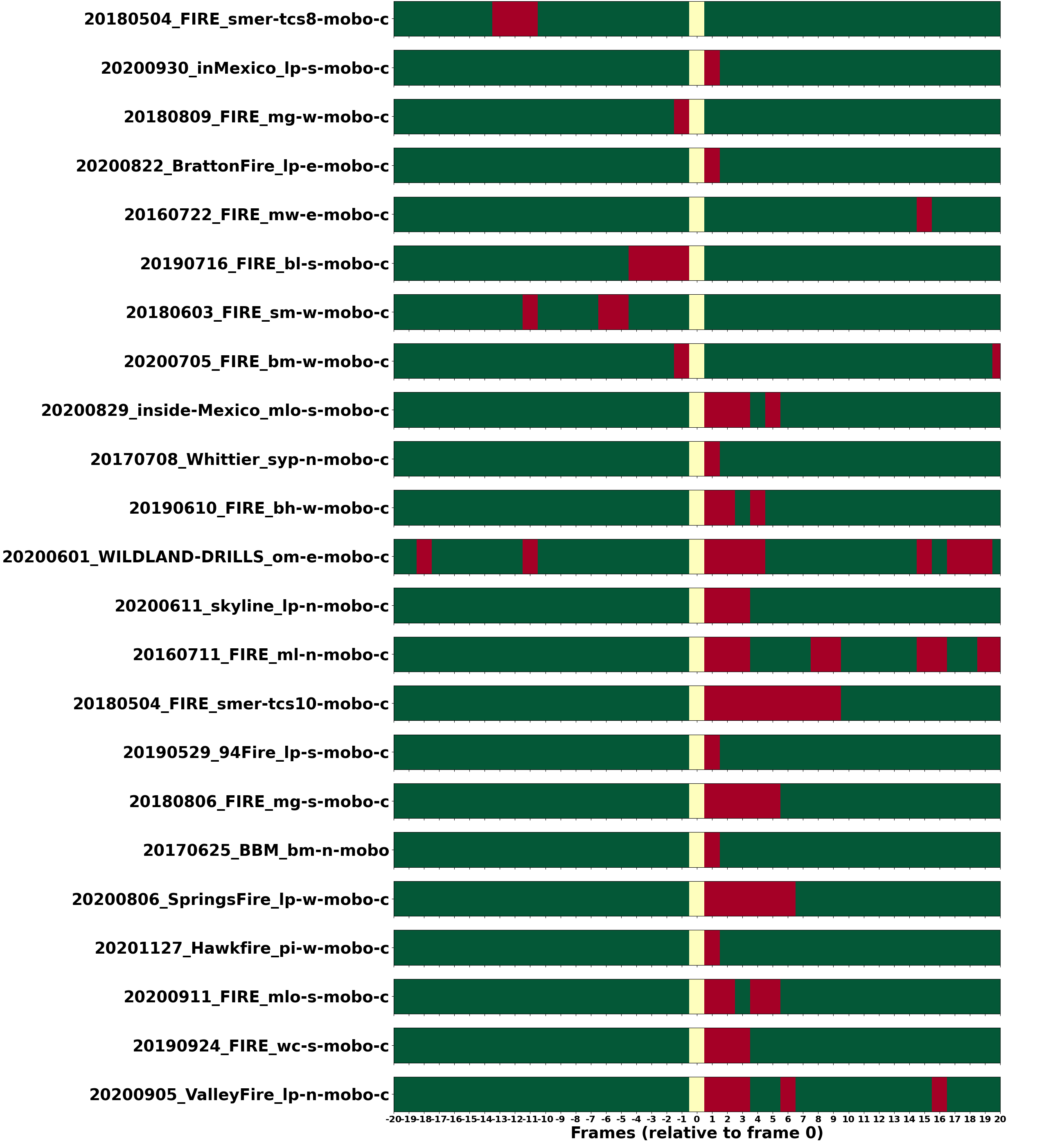} 
        \caption{Single-frame model prediction compared to the ground-truth}
    \end{subfigure}
    \hspace{0.05\textwidth} 
    \begin{subfigure}[b]{0.45\textwidth}
        \centering
        \includegraphics[width=\textwidth]{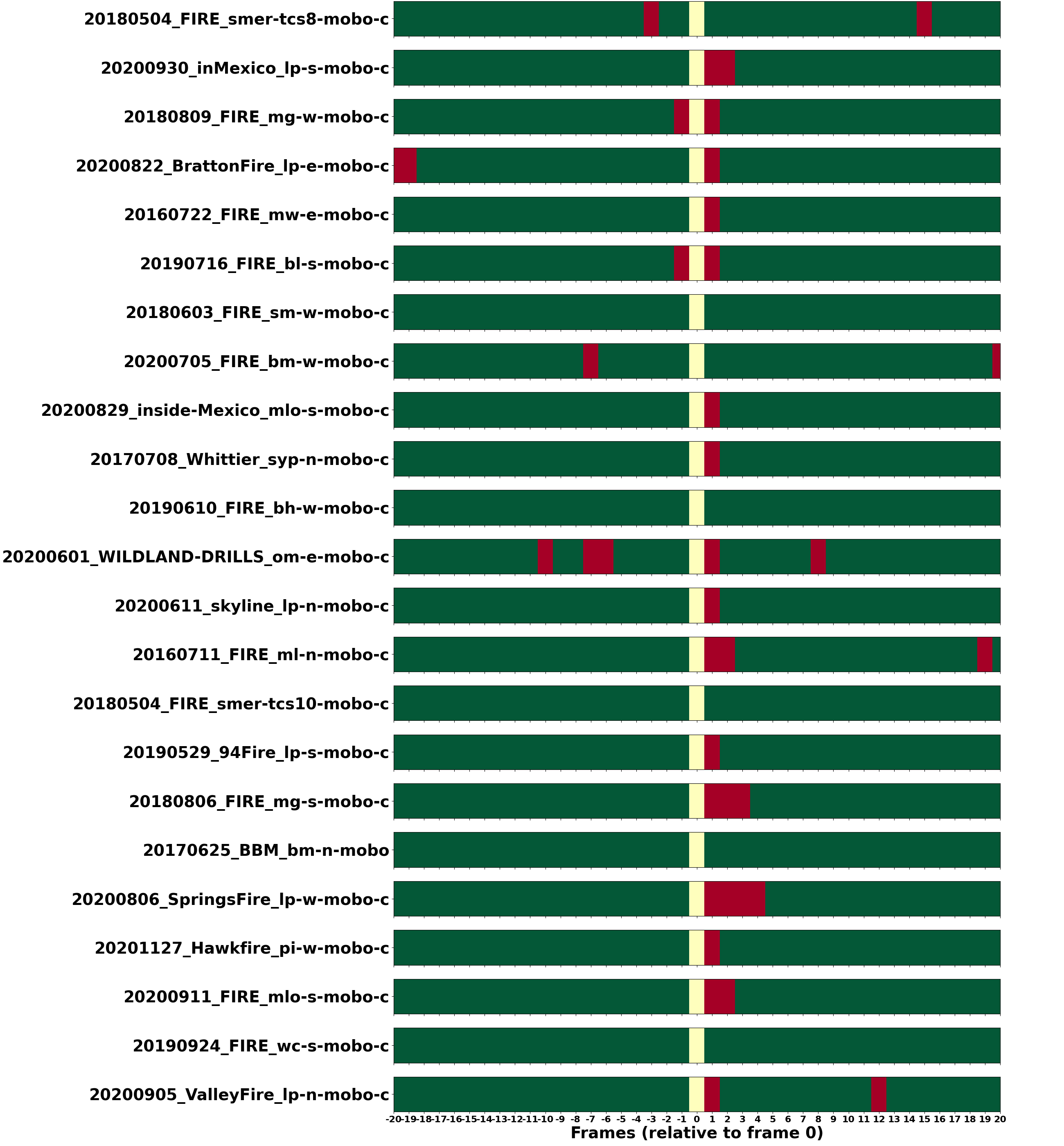} 
        \caption{Video model prediction compared to the ground-truth}
    \end{subfigure}
    \caption{Comparison of the predictions of a single-frame model versus a video model on a set of videos from FigLib.} \label{fig:video_prediction}
\end{figure*}

Table \ref{app:tab_elapsed} shows the time elapsed from the start of the fire until the first correct detection of smoke. The average mean time for first detection over the whole test set dataset is 19 seconds earlier (1.24 vs 0.94 mn). 
More fine-grained visualization of the predictions are available in Figure \ref{fig:video_prediction}, which depicts for a set a videos if the predictions are true or false at each frame. It is possible to see that the sequential model allows for detecting wildfire earlier in general (12 vs 4 videos where sequential detects before than Vanilla), and also for smoothing the prediction (false negative after a detection). 

\begin{table*}[ht]
\centering
\resizebox{.7\textwidth}{!}{
\begin{tabular}{@{}lcc@{}}
\toprule
\multicolumn{1}{c}{Video Name} & \multicolumn{2}{c}{Time Elapsed (min)} \\
\midrule
& YOLOv8 & YOLOv8 + CNN-LSTM \\
\midrule
20201208\_FIRE\_om-s-mobo-c             & 3 & 2 \\
20190716\_FIRE\_bl-s-mobo-c             & 0 & 1 \\
20160722\_FIRE\_mw-e-mobo-c             & 0 & 1 \\
20180726\_FIRE\_so-w-mobo-c             & 1 & 1 \\
20180806\_FIRE\_mg-s-mobo-c             & 5 & 3 \\
20180809\_FIRE\_mg-w-mobo-c             & 0 & 1 \\
20170708\_Whittier\_syp-n-mobo-c        & 1 & 1 \\
20190529\_94Fire\_lp-s-mobo-c           & 1 & 1 \\
20200806\_SpringsFire\_lp-w-mobo-c      & 6 & 4 \\
20200822\_BrattonFire\_lp-e-mobo-c      & 1 & 1 \\
20200930\_inMexico\_lp-s-mobo-c         & 1 & 2 \\
20170625\_BBM\_bm-n-mobo                & 0 & 0 \\
pyronear\_st\_peray\_1                  & 0 & 0 \\
cabanelle-125\_2024-04-03T10-16-30      & 1 & 1 \\
cabanelle-327\_2024-02-27T08-20-51      & 1 & 1 \\
cabanelle-125\_2024-02-27T14-33-57      & 1 & 1 \\
cabanelle-244\_2024-01-01T11-07-26      & 1 & 1 \\
cabanelle-327\_2024-02-26T14-48-09      & 0 & 0 \\
cabanelle-244\_2024-04-07T09-28-19      & 1 & 0 \\
cabanelle-244\_2024-02-27T13-07-58      & 4 & 2 \\
cabanelle-244\_2024-02-23T10-22-27      & 2 & 1 \\
cabanelle-125\_2024-04-10T08-18-37      & 0 & 0 \\
cabanelle-244\_2024-01-04T14-31-58      & 0 & 0 \\
cabanelle-327\_2024-02-27T09-08-53      & 0 & 0 \\
cabanelle-125\_2024-04-03T09-01-27      & 2 & 0 \\
awf\_nvseismolab\_noaaX-0180            & 1 & 1 \\
awf\_nvseismolab\_peavineX00056         & 2 & 2 \\
awf\_nvseismolab\_geyserpeakX-0098      & 0 & 0 \\
2025-01-28T11-28-50\_camera:\_gupo-0347 & 1 & 1 \\
2025-01-30T20-14-15\_camera:\_gupo-0347 & 2 & 0 \\
2025-01-30T00-05-39\_camera:\_gupo-0347 & 0 & 0 \\
2025-01-30T01-12-41\_camera:\_gupo-0347 & 0 & 0 \\
2025-01-30T20-46-03\_camera:\_gupo-0347 & 2 & 2 \\
ADF\_1320                                & 2 & 1 \\
\midrule
Mean $\pm$ sd                              & (1.24 $\pm$ 1.44) & (0.94 $\pm$ 0.938) \\
+ 64 samples                           & (1.76  $\pm$ 1.53) & (1.17 $\pm$ 1.37)        \\
\bottomrule
\end{tabular}
}
\caption{Time elapsed before detecting the fire, using YOLOv8 (one frame) and YOLOv8+CNN-LSTM. The top section reports results on the subsampling shown; the bottom row (“+ 64 samples”) gives the aggregate results for the full dataset.}
\vspace*{-.2cm}
\label{app:tab_elapsed}
\end{table*}

\section{SmokeFrames-50k annotations} \label{app:early_detection}

When looking at the ground truth annotations of SmokeFrames-50k, we realized that there were many frames in which our model detected a wildfire when there was no annotation. 
This was due to the camera moving to center around the smoke plume, or because the wildfire was annotated with a bit of latency. 

\section{Other Datasets}

We also reviewed other published datasets in this work, but decided to discard them after manual reviewing because of the nature of the images, with a distribution very different: huge wildfires, or within city environment \cite{Casas2023,DeVenancio2022}.

\end{document}